\newcount\Comments  
\documentclass[11pt,a4paper]{article}
\usepackage[hyperref]{acl2020}
\usepackage{times}
\usepackage{latexsym}

\usepackage{microtype}

\usepackage{graphicx}
\usepackage{xspace}
\usepackage{tabularx}
\usepackage{booktabs}
\usepackage{multirow}
\usepackage{comment}
\usepackage{caption}
\usepackage{subcaption}
\usepackage{textgreek}
\usepackage{amssymb}
\usepackage{amsmath}
\usepackage{color, colortbl}
\usepackage{newunicodechar}
\usepackage{tipa}
\usepackage[T1]{fontenc} 
\usepackage{enumerate}

\usepackage[russian,english]{babel}
\usepackage{siunitx}
\usepackage{pifont} 
\usepackage{color}
\newcommand{\kibitz}[2]{\ifnum\Comments=1\textcolor{#1}{#2}\fi}

\aclfinalcopy 

\usepackage{cleveref}
\usepackage{multirow, makecell}

\usepackage{xurl}

\title{Separating Grains from the Chaff: Using Data Filtering to Improve Multilingual Translation for Low-Resourced African Languages}

\author{\normalsize Idris Abdulmumin$^{1,2\thanks{* Equal contribution.}}$, Michael Beukman$^{3\footnotemark[1]}$, Jesujoba O. Alabi$^{4\footnotemark[1]}$,  Chris Emezue$^{5,6}$,\\
\textbf{\normalsize Everlyn Asiko$^{7,8}$, Tosin Adewumi$^{9}$, Shamsuddeen Hassan Muhammad$^{2,10}$,} \\
\textbf{\normalsize Mofetoluwa Adeyemi, Oreen Yousuf$^{11}$, Sahib Singh$^{12}$, Tajuddeen Rabiu Gwadabe$^{2,13}$\medskip} \\
\footnotesize All: MasakhaneNLP, $^{1}$Ahmadu Bello University, Zaria, Nigeria, $^{2}$HausaNLP,  $^{3}$University of the Witwatersrand, South Africa, \\
\footnotesize $^{4}$Saarland University, Germany, $^{5}$TUM, Germany, $^{6}$Mila - Quebec AI Institute, $^{7}$University of Cape Town, South Africa,\\
\footnotesize $^{8}$African Institute for Mathematical Sciences, $^{9}$Luleå University of Technology, Sweden, $^{10}$LIAAD-INESC TEC, Porto, Portugal,\\
\footnotesize $^{11}$Uppsala University, Sweden, $^{12}$Ford Motor Company, $^{13}$University of Chinese Academy of Sciences, China \\
\footnotesize \texttt{iabdulmumin@abu.edu.ng}\\

}
\date{}

\begin{document}
\maketitle
\begin{abstract}
We participated in the WMT 2022 Large-Scale Machine Translation Evaluation for the African Languages Shared Task. This work describes our approach, which is based on filtering the given noisy data using a sentence-pair classifier that was built by fine-tuning a pre-trained language model. To train the classifier, we obtain positive samples (i.e. high-quality parallel sentences) from a gold-standard curated dataset and extract negative samples (i.e. low-quality parallel sentences) from automatically aligned parallel data by choosing sentences with low alignment scores. Our final machine translation model was then trained on filtered data, instead of the entire noisy dataset. We empirically validate our approach by evaluating on two common datasets and show that data filtering generally improves overall translation quality, in some cases even significantly.
\end{abstract}

\section{Introduction}
This paper presents Masakhane NLP's submission to the WMT 2022 large-scale machine translation evaluation for African languages. We participated in the constrained translation task and chose to focus on a subset of all the language pairs considered for this task due to resource constraints. We specifically explore the language directions \textbf{\{\texttt{hau}, \texttt{ibo}, \texttt{lug}, \texttt{swa}, \texttt{tsn}, \texttt{yor}, \texttt{zul}\}$\leftrightarrow$\texttt{eng}} and \textbf{\texttt{wol}$\leftrightarrow$\texttt{fra}}, and submitted our primary and secondary systems which were competitive with other submissions for this task.

Machine translation has received much attention recently, especially for low-resourced languages~\citep{adelani-etal-2022-thousand,fanetlm2m100,haddow2021Survey,hoang-etal-2018-iterative,nekoto2020Participatory}. A promising approach for such setups is to fine-tune large pre-trained language models on the available small amount of translation data~\citep{neubig-hu-2018-rapid,adelani-etal-2021-effect,adelani-etal-2022-thousand}. While most of these language models are trained on predominantly high-resourced language datasets~\citep{conneau2019Unsupervised,devlin-etal-2019-bert,radford2018Improving}, there have been a few models that were pre-trained~\citep{ogueji2021Small} or adaptively fine-tuned~\citep{afro_maft} only on low-resourced languages.

Recent works have tried, successfully, to supplement the existing small amounts of natural data in low-resource languages with artificially generated parallel data. For instance, in machine translation, \citet{sennrich-etal-2016-improving} and \citet{Ueffing2006} padded the true parallel data with automatic translations of monolingual sentences through back-translation and self-learning respectively. Others, such as \citet{banon-etal-2020-paracrawl,elkishky_ccaligned_2020}; and \citet{schwenk-etal-2021-ccmatrix}, have used different approaches for detecting potentially aligned sentences within web datasets. While significant improvements have been achieved with these synthetic datasets, an in-depth investigation by \citet{10.1162/tacl_a_00447} has found them to be fraught with many issues, such as misalignment, wrongful language codes, etc.

Similarly, research has shown that data quality plays an important role in the performance of natural language processing (NLP) models, in machine translation specifically~\citep{arora2021Studying,dutta-etal-2020-uds,hasan2020Lowresource,tchistiakova-etal-2021-edinsaar}, but also more generally in other NLP tasks~\citep{abdul2012Extrinsic,alabi2019Massive}. It was found that often times, models that were trained on smaller amounts of high-quality data outperform their counterparts that are trained on larger amounts of noisy datasets \citep{Gasco2012,przystupa-abdul-mageed-2019-neural,abdulmumin2022quantity,gibert-etal-2022-quality}. This has led to many studies~\citep{eetemadi2015Survey} and prior WMT tasks~\citep{koehn-etal-2018-findings,koehn-etal-2019-findings,koehn-etal-2020-findings} that attempt to find ways to improve the quality of existing data, which, as mentioned before, is often rife with errors.

Therefore, in our submission to the shared task, we experimented with filtering web-mined data for African languages using pre-trained language models and evaluated the effect of using this filtered data on machine translation performance. We defined our filtering approach as a sentence-pair binary classification task and fine-tuned a pre-trained language model using positive and negative samples. We used sentences from the high-quality MAFAND-MT~\citep{adelani-etal-2022-thousand} dataset (which was included in the training data for the constrained task) as positive examples and created negative examples by extracting sentences with low language-agnostic sentence representations (LASER)~\citep{artetxe2018Massively} alignment scores from the \texttt{wmt22-african}~\citep{nllb2022} corpus that was provided for this task. Our results highlight the importance of filtering on the quality of the final machine translation system. We also detail how to create a high-quality filter for African languages using a few gold-standard parallel sentences. We release our codes on GitHub.\footnote{\url{https://github.com/abumafrim/WMT22-MasaKhane}}

The rest of the paper is organized as follows: in Section \ref{sec:related}, we review related work, and in Section \ref{sec:dataset}, we present the dataset we used. Section \ref{sec:filtering} provides an overview of the bitext filtering approach, while Section \ref{sec:experiment} details experimental settings and the translation model architecture. In Section \ref{sec:result}, we evaluate the model's performance, and lastly in Section \ref{sec:conclude}, we conclude and highlight some future research directions.

\section{Related Work} \label{sec:related}

One of the difficulties when dealing with low-resourced settings, as we do here, is that high-quality parallel texts are particularly scarce~\cite{koehn-knowles-2017-six}. To curate data for such language pairs, methods for automatically mining parallel text from the web using heuristics \cite{resnik-1999-mining} or latent space and similarity-based filters \cite{artetxe-schwenk-2019-margin, schwenk-etal-2021-ccmatrix} have been proposed. These have led to the curation of publicly available web-mined datasets such as CCAligned \cite{elkishky_ccaligned_2020}, CCMatrix \cite{fanetlm2m100,schwenk-etal-2021-ccmatrix}, ParaCrawl \cite{espla-etal-2019-paracrawl}, and WikiMatrix \cite{WikiMatrix} to mention just a few.

However, the recent research work by \citet{10.1162/tacl_a_00447} shows that the automatically aligned and mined parallel bitexts, especially for low-resource language pairs, contain various degrees of errors and less than half of the data are of good quality. 
Additionally, many approaches generate large amounts of synthetic data, often through back-translation, where synthetic parallel data is generated by automatically translating monolingual data~\citep{bojar2011Improving,lambert2011Investigations,sennrich-etal-2016-improving}. While additional data has the potential to improve the trained models, these synthetic datasets are often of low quality~\citep{xu2019Improving}.
These observations have led to an increased interest in the automatic filtering of noisy bitexts as a key research topic in machine translation (MT). 

\begin{table*}
    \small
    \centering
    \begin{tabular}{@{}lp{10cm}p{3.5cm}@{}}\toprule
        Direction & Parallel sentences & Problem \\ \midrule
         \texttt{eng} $\rightarrow$ \texttt{hau} & \makecell[tl]{\textbf{src:} I booked the house for my husband's family as we were get-\\ting married in Ericeira.\\\textbf{tgt:} na tsarr da aba a ka kasarr ni ila ure imbarr yi ngbangbamu.} & \textbf{tgt} is not a Hausa sentence \\
         \texttt{eng} $\rightarrow$ \texttt{hau} & \makecell[tl]{\textbf{src:} "Go hunt, and may the light be with you."""\\\textbf{tgt:} """Zo, zo muje, ke kika hada fitinar ke za ki warware ta."""} & \textbf{tgt} is not a translation of the \textbf{src} \\
         \texttt{eng} $\rightarrow$ \texttt{hau} & \makecell[tl]{\textbf{src:} The Moslem creed.\\\textbf{tgt:} Musa Aminta} & mismatched named entities \\
         \texttt{eng} $\rightarrow$ \texttt{hau} & \makecell[tl]{\textbf{src:} Israel\\\textbf{tgt:} \foreignlanguage{russian}{оооооооооооооооооооооооооооооооооооооооооооооооо}\\ \foreignlanguage{russian}{оооооооовввввввввввввввввввввввв}} & mistranslation; foreign characters \\\bottomrule
    \end{tabular}
    \caption{Examples of noise in the auto-aligned bitext}
    \label{tab:noise}
\end{table*}

One approach to improve data quality is to filter out the noisy or invalid parts of a large corpus, keeping only a high-quality subset thereof \citep{abdulmumin-hybrid-2021}. In this vein, numerous filtering methods have been developed~\citep{axelrod2011Domain, eetemadi-toutanova-2015-detecting,junczysdowmunt2018Dual}. For instance, \citet{xu2019Improving} use the cosine similarity between sentence embeddings as a measure of how closely aligned two sentences are. \citet{imankulova2017Improving} perform back-translation and then filter based on the sentence-level BLEU score, keeping only those sentences with a high BLEU. Similarly, \citet{adjeisah2021Pseudotext} perform a round-trip translation and only use the sentence pair if it is sufficiently close to the original sentence, according to a chosen similarity measure. There has also been work on alignment between two parallel corpora, and \citet{hasan2020Lowresource} uses the LASER score\footnote{\url{https://github.com/facebookresearch/LASER}} to evaluate alignment, and filter out all sentences below a specific threshold.

\section{Datasets} \label{sec:dataset}
We participated in the constrained translation track and used only the provided dataset. We present the various dataset used, their sizes and corresponding sources in~\Cref{tab:datastat} in~\Cref{app:data-sources}. For our experiment, we selected 8 language pairs and developed different multilingual machine translation systems for them. These language pairs are  \textbf{\{\texttt{hau}, \texttt{ibo}, \texttt{lug}, \texttt{swa}, \texttt{tsn}, \texttt{yor}, \texttt{zul}\}$\leftrightarrow$\texttt{eng}} and \textbf{\texttt{wol}$\leftrightarrow$\texttt{fra}}. According to the recommendation of \citet{10.1162/tacl_a_00447}, we carefully examined the training dataset provided by manual inspection and divided it into two categories based on the source of the data and the amount of noise included therein. In the following subsections, we describe these two categories of data.

\subsection{Clean Bitext}
\label{sec:clean}
This category of training data comprises all the datasets that are considered to be manually curated.
The datasets in this category include: bible-uedin~\cite{Christodouloupoulos2015}, MAFAND-MT,\footnote{\url{https://github.com/masakhane-io/lafand-mt.git}} QED~\cite{abdelali-etal-2014-amara}, Mozilla-I10n,\footnote{\url{https://github.com/mozilla-l10n/mt-training-data}} Tanzil,\footnote{\url{https://tanzil.net/trans/}} and several others listed in \Cref{tab:datastat}. The clean bitext consists of sentences mostly in the news and religious domains, with a few in the health, education, and technology domains. We also refer to the clean bitext as True Parallel in this paper.

\subsection{Noisy Bitext}
\label{sec:noisy}

We categorized all the automatically aligned bitext as noisy bitext. This also includes the LASER filtered data. The sentences in this category make up the majority of the training dataset, making up 99.2\% of the total training data. The datasets in this category include: CCAligned, CCMatrix, LASER \texttt{wmt22\_african},\footnote{\url{https://huggingface.co/datasets/allenai/wmt22_african}} WebCrawl African,\footnote{\url{https://github.com/pavanpankaj/Web-Crawl-African}} and the following datasets from OPUS \cite{tiedemann-2012-parallel}: MultiCCAligned \cite{elkishky_ccaligned_2020}, TED2020 \cite{reimers-2020-multilingual-sentence-bert}, WikiMatrix \cite{WikiMatrix}, XLEnt \cite{elkishky_xlent_2021} and others highlighted in \Cref{tab:datastat}.

\begin{table}[t]
    \centering
    \scalebox{0.85}{
    \small
    \begin{tabular}{rrrr}\toprule
    \multicolumn{2}{l}{Language pair} & Data size & \% of original \\\midrule
        \texttt{eng} & \texttt{hau} & $9,122,559$ & $99.9$ \\
        & \texttt{ibo} & 520,544 & $99.6$ \\
        & \texttt{lug} & 3,511,275 & $99.8$ \\
        & \texttt{swa} & 32,898,533 & $99.6$ \\
        & \texttt{tsn} & 6,036,656 & $99.1$ \\
        & \texttt{yor} & 1,718,105 & $99.3$ \\
        & \texttt{zul} & 4,142,146 & $97.6$ \\
        \texttt{fra} & \texttt{wol} & 237,348 & $100.0$ \\\bottomrule
    \end{tabular}
    }
    \caption{Training data after filtering using heuristics}
    \label{tab:data-heuristics}
\end{table}

On manual inspection, however, we found numerous issues with the data, including non-parallel sentences, sentences that consist of only numbers and/or punctuation, sentences in different languages, etc. Examples of noise in the auto-aligned data can be seen in \Cref{tab:noise}.

\subsection{Validation and Test Data}\label{sec:vald}
For the optimization of our translation systems, we combined the FLORES-101~\citep{10.1162/tacl_a_00474} and MAFAND-MT~\citep{adelani-etal-2022-thousand} development sets for each of the 8 language pairs. To compare the performance of the developed MT engines, we evaluated on the FLORES-101 devtest set and the MAFAND-MT test set.

\section{Parallel Data Filtering} \label{sec:filtering}
To attempt to deal with the highly noisy data, we opted to use filtering techniques to remove many invalid or incorrectly aligned sentences, similar to prior work~\citep{arora2021Studying,hasan2020Lowresource,xu2019Improving}. We first used some simple heuristic approaches, described in \Cref{sec:filter:heuristic}, and then progress to an automatic filtering method, detailed in \Cref{sec:filter:auto}.

\subsection{Heuristics}\label{sec:filter:heuristic}
We filtered sentences that consist of only numbers and/or punctuation marks. After filtering, the statistics of the resulting training dataset are shown in \Cref{tab:data-heuristics}. The table shows that $2.4\%$ of the original Zulu (\texttt{zu}) data consisted of just numbers or punctuation, while other languages had smaller invalid portions, between $0.0\%$ and $0.1\%$.

\subsection{Automatic Filtering}\label{sec:filter:auto}
Due to the large size of the automatically aligned dataset, we adopted an automatic approach to determine the quality of parallel sentences to train our translation models. The approach we adopted is sentence-pair binary classification \cite{nguyen-etal-2021-combining}, where we used a transformer-based model to predict the probability that two aligned sentences are actual translations of each other.
We explain the process of training data generation and the experimental choices for building the filtering model.

\begin{table}[t]
    \centering
    \scalebox{0.60}{
    \begin{tabular}{lllllllll}\toprule
    \multirow{2}{*}{Data} & \multicolumn{7}{c}{eng} & \multicolumn{1}{c}{fra} \\\cmidrule(lr){2-8}\cmidrule(lr){9-9}
        & \texttt{hau}$^\dagger$ & \texttt{ibo}$^\dagger$ & \texttt{lug}$^\dagger$ & \texttt{swa}$^\dagger$ & \texttt{tsn}$^\dagger$ & \texttt{yor}$^\dagger$ & \texttt{zul}$^\dagger$ & \texttt{wol}$^\dagger$ \\
    \midrule
        Train & $6,198$ & $13,998$ & $8,152$ & $61,566$ & $4,202$ & $13,290$ & $7,002$ & $6,722$ \\
        Dev & $2,602$ & $3,002$ & $3,002$ & $3,584$ & $2,686$ & $3,090$ & $2,480$ & $3,014$ \\
        Test & $3,002$ & $3,002$ & $3,002$ & $3,672$ & $3,002$ & $3,118$ & $1,998$ & $3,002$ \\\bottomrule
    \end{tabular}
    }
    \vspace{-3mm}
    \caption{Sentence-pair classification training data: a mixture of MAFAND-MT$^\dagger$ sentence pairs, taken as positive samples, and \texttt{wmt22\_african} (worst pairs based on LASER scores), taken as negative samples.}
    \label{tab:spc_training_data}
\end{table}

\begin{table*}[t]
    \scalebox{0.75}{
    \begin{tabular}{llrrrrrrrrr}\toprule
        \multirow{2}{*}{Model} & & \multicolumn{7}{c}{en} & \multicolumn{1}{c}{fr} & \multirow{2}{*}{\textbf{$F1_{avg.}$}} \\ \cmidrule(lr){3-9} \cmidrule(lr){10-10}
        & & \texttt{hau} & \texttt{ibo} & \texttt{lug} & \texttt{swa} & \texttt{tsn} & \texttt{yor} & \texttt{zul} & \texttt{wol} & \\ \midrule
        ALBERT-base & F1 & $95.6$ & $94.2$ & $94.7$ & $89.6$ & $95.7$ & $91.1$ & $87.4$ & $95.1$ & $92.9$ \\
        & \texttt{t=0.5} & $278,930$ & $78,056$ & $119,516$ & $5,832,820$ & $346,329$ & $151,886$ & $363,739$ & $6,552$ & \\
        & \texttt{t=0.7} & $197,232$ & $63,207$ & $82,243$ & $3,921,959$ & $252,499$ & $91,366$ & $213,991$ & $4,365$ & \\ \midrule
        ALBERT-xlarge & F1 & $93.2$ & $92.8$ & $96.3$ & $63.7$ & $95.3$ & $90.7$ & $89.1$ & $84.4$ & 88.2 \\
        & \texttt{t=0.5} & $115,987$ & $129,304$ & $146,948$ & $3,263,429$ & $273,154$ & $113,860$ & $613,483$ & $49,926$ & \\
        & \texttt{t=0.7} & $81,641$ & $111,562$ & $102,354$ & $1,638,528$ & $217,200$ & $86,558$ & $302,951$ & $41,283$ & \\ \midrule
        AfroXLMR-base & F1 & $96.9$ & $94.4$ & $95.4$ & $94.6$ & $96.1$ & $98.4$ & $88.0$ & $97.1$ & $95.1$\\
        & \texttt{t=0.5} & $296,881$ & $75,102$ & $149,051$ & $6,139,327$ & $363,155$ & $81,902$ & $281,803$ & $6,997$ & \\
        & \texttt{t=0.7} & $226,666$ & $59,995$ & $84,499$ & $5,064,365$ & $276,490$ & $73,786$ & $171,778$ & $5,189$ & \\ \bottomrule
    \end{tabular}
    }
    \caption{Training data after filtering using sentence-pair classifier --- t=Threshold; F1 was computed at \texttt{t=0.5}}
    \label{tab:data-spc}
\end{table*}

\subsubsection{Positive and negative samples}

To create the training and evaluation data for the sentence-pair classification-based filtering, we generated positive and negative samples from the training data available for this task. We used the train, dev and test sets from the MAFAND-MT dataset, which is a gold-standard parallel dataset, as positive examples. For the negative examples, however, we sorted the sentences in \texttt{wmt22\_african} dataset that was provided for this task based on their LASER alignment scores, and selected the least scored sentences in equal amounts to each of the positive examples. The distribution of the train, dev and test samples is presented in \Cref{tab:spc_training_data}.

\subsubsection{Model}
We fine-tuned two pre-trained language models, ALBERT \cite{Lan2020ALBERT:} and AfroXLMR \cite{afro_maft} for the sentence pair binary classification task. ALBERT was selected based on its performance on downstream NLP tasks \cite{Lan2020ALBERT:}, even though it has fewer parameters than other BERT-based models \cite{nguyen-etal-2021-combining}. AfroXLMR, on the other hand, was chosen because it was trained on African languages \cite{afro_maft}, and such a setup has been shown to improve performance on downstream tasks for these languages \cite{adelani-etal-2022-thousand}.

\subsection{Filter Training Setup}

The filtering models were trained to accept a pair of sentences from the source and target languages. During training, the \texttt{[CLS]} token hidden representation of the input sentence pairs is fed into a linear Layer and the model is optimized using binary cross entropy loss. However, at inference time, we add a sigmoid layer to the output to predict a number between $0.0$ and $1.0$ indicating the likelihood of the bitexts being translations of each other. We fine-tuned these models using each language's train split of positive and negative samples, then evaluated performance on the test set while optimizing on the development set.

The performance of the various automatic filtering models and the subsequent sizes of the filtered datasets for the 8 language pairs are shown in \Cref{tab:data-spc}. This table shows the number of sentence pairs the models classified as actual translation pairs using a threshold of $0.5$ and $0.7$ as well as the F1 score when using the $0.5$ threshold. Additionally, in \Cref{tab:data-overlap}, we show the number of sentences that were classified by two or all three of the models as being high-quality.

\begin{table}[t]
    \scalebox{0.8}{
    \begin{tabular}{lrrr}\toprule
        \texttt{t=0.5} & Albert-base & Albert-xlarge & AfroXLMR \\ \midrule
        Albert-base & 2,984,862 & 1,750,707 & 2,575,408 \\
        Albert-xlarge & - & 2,107,204 & 1,058,711 \\
        AfroXLMR & - & - & 3,925,612 \\ \midrule
        \textbf{sents. in ALL} & & & \textbf{668,633} \\ \toprule
        \texttt{t=0.7} & & & \\ \midrule
        Albert-base & 1,977,486 & 909,203 & 1,884,922 \\
        Albert-xlarge & - & 1,206,493 & 547,925 \\
        AfroXLMR & - & - & 3,420,147 \\ \midrule
        \textbf{sents. in ALL} & & & \textbf{331,208} \\ \bottomrule
    \end{tabular}
    }
    \caption{Data overlap after filtering using the sentence-pair classifier models}
    \label{tab:data-overlap}
\end{table}

\section{MT Experiments} \label{sec:experiment}
To evaluate the effect of our filtering techniques, we trained some multilingual NMT models for the 8 language pairs that we have selected for this task. In the following subsections, we highlight the model architectures, training setups, and different multilingual models that were trained.

\subsection{Model Architecture}
For our experiments, we fine-tune M2M-100 ~\cite{fanetlm2m100} on different subsets of the provided data. M2M-100 is a pretrained translation model trained on several languages including African languages, as such it has seen all the languages we have chosen for this task during pre-training. We  use the model with $418M$ parameters.

\begin{table*}
 \begin{center}
 \resizebox{\textwidth}{!}{
  \begin{tabular}{lrrrrrrrrrrrrrrrrr}\toprule
    \multirow{2}{*}{Models} & \multicolumn{7}{c}{\texttt{eng}$\rightarrow$x} & \multicolumn{1}{c}{\texttt{fra}$\rightarrow$x} & \multicolumn{7}{c}{x$\rightarrow$\texttt{\textit{eng}}} & \multicolumn{1}{c}{x$\rightarrow$\texttt{\textit{fra}}} & \multirow{2}{*}{Avg.} \\ 
    \cmidrule(lr){2-8}\cmidrule(lr){9-9}\cmidrule(lr){10-16}\cmidrule(lr){17-17}
     & \texttt{hau} & \texttt{ibo} & \texttt{lug} & \texttt{swa} & \texttt{tsn} & \texttt{yor} & \texttt{zul} & \texttt{wol} & \texttt{hau} & \texttt{ibo} & \texttt{lug} & \texttt{swa} & \texttt{tsn} & \texttt{yor} & \texttt{zul} & \texttt{wol} & \\ \midrule
    \multicolumn{18}{l}{\textbf{BLEU}} \\
    \midrule
    \multicolumn{18}{l}{\textbf{~~Baselines}} \\
    ~~Clean bitext & 9.30 & 13.19 & \textbf{4.00} & 23.17 & 8.56 & \textbf{3.60} & 9.43 & \textbf{3.56} & 14.24 & 12.56 & 11.24 & 26.86 & 8.78 & 8.90 & 18.51 & 6.03 & 11.37 \\
    ~~Noisy bitext & 15.32 & 10.77 & 2.14 & 30.64 & 12.87 & 2.57 & 12.35 & 0.69 & 20.58 & 14.69 & 13.19 & 31.80 & 16.29 & 11.40 & 24.68 & 3.22 & 13.95 \\
    ~~Clean $+$ Noisy bitext & 15.34 & 11.37 & 2.40 & 30.48 & 13.31 & 2.48 & 12.61 & 0.73 & 20.53 & 15.07 & 13.34 & 31.61 & 16.50 & 11.75 & 24.29 & 3.88 & 14.11 \\ \midrule
    \multicolumn{18}{l}{\textbf{~~Filtered only}} \\
    ~~albert-xlarge-0-7 & 16.43 & 15.38 & 2.54 & 29.89 & \textbf{16.31} & 3.00 & 15.18 & 0.65 & 20.05 & 17.32 & 12.51 & 34.24 & 18.55 & 12.62 & 27.31 & 5.14 & 15.45 \\
    \midrule
    \multicolumn{18}{l}{\textbf{~~Filtered $+$ Clean bitext}} \\
    ~~albert-xlarge-0-5 & 16.05 & 15.01 & 3.22 & \textbf{33.31} & 15.96 & 3.08 & 14.97 & 1.99 & \textbf{20.92} & 17.45 & 13.93 & \textbf{34.99} & 18.24 & 13.24 & 27.65 & 6.43 & 16.03 \\
    ~~albert-xlarge-0-7 & 16.55 & \textbf{15.70} & 3.45 & 31.97 & \textbf{16.31} & 3.16 & \textbf{15.50} & 2.12 & 20.85 & \textbf{17.88} & \textbf{13.97} & 34.40 & \textbf{18.29} & \textbf{13.38} & \textbf{27.35} & \textbf{7.20} & \textbf{16.13} \\ \midrule
    \multicolumn{18}{l}{\textbf{\textsc{chrF}}} \\ \midrule
    \multicolumn{18}{l}{\textbf{~~Baselines}} \\
    ~~Clean bitext & 34.01 & \textbf{45.31} & \textbf{42.14} & 55.14 & 45.58 & \textbf{30.56} & 43.62 & \textbf{30.55} & 34.70 & \textbf{45.20} & \textbf{46.21} & 54.53 & 45.37 & 39.04 & 46.50 & \textbf{30.77} & 41.83 \\
    ~~Noisy bitext & 30.04 & 34.18 & 33.04 & 54.34 & 43.51 & 16.23 & 46.30 & 8.92 & 35.15 & 37.46 & 35.96 & 54.38 & 45.39 & 33.84 & 49.85 & 15.72 & 35.89 \\
    ~~Clean $+$ Noisy bitext & 30.53 & 35.75 & 33.69 & 54.66 & 44.23 & 15.90 & 46.37 & 10.91 & 35.70 & 38.82 & 37.50 & 54.78 & 45.62 & 35.35 & 49.71 & 19.21 & 36.79 \\
    \midrule
    \multicolumn{18}{l}{\textbf{~~Filtered only}} \\
    ~~albert-xlarge-0-7 & 36.18 & 41.71 & 36.94 & 54.79 & 51.64 & 18.85 & 51.14 & 10.86 & 37.18 & 41.38 & 39.07 & 56.81 & \textbf{56.81} & 38.27 & 52.71 & 22.20 & 40.41 \\ \midrule
    \multicolumn{18}{l}{\textbf{~~Filtered $+$ Clean bitext}} \\
    ~~albert-xlarge-0-5 & 36.56 & 43.19 & 40.44 & 56.65 & 51.25 & 20.33 & 50.77 & 23.44 & 37.79 & 43.72 & 44.34 & 57.70 & 51.98 & 40.01 & 52.75 & 27.76 & 42.42 \\
    ~~albert-xlarge-0-7 & 36.64 & 44.32 & 41.44 & 56.60 & \textbf{52.98} & 21.88 & \textbf{51.43} & 25.22 & 38.11 & 44.23 & 45.14 & \textbf{57.73} & 52.22 & \textbf{40.68} & \textbf{53.02} & 29.29 & \textbf{43.18} \\
    \bottomrule
    \end{tabular}
  }
  \vspace{-1mm}
  \caption{Performance of the multilingual model on the \textbf{FLORES-101} devtest set, with the maximum BLEU per column in \textbf{bold}. x represents African languages.}
  \label{tab:flores}
\end{center}
\end{table*}

\subsection{Training Setup}
We fine-tuned the M2M-100 model based on the implementation within the Fairseq\footnote{\url{https://github.com/facebookresearch/fairseq}} toolkit \cite{ott-etal-2019-fairseq}. We used batch sizes of $2,048$ tokens, a maximum sentence length of $1,024$, and a dropout of $0.3$. For optimization, we used Adam \cite{kingma:adam} with \(\beta_1\) = $0.9$ and \(\beta_2\) = $0.998$, a learning rate of $5e-5$ and a warmup of $2,500$ updates. The optimizer uses a label-smoothed cross-entropy loss function with a label-smoothing value of $0.2$. All models were trained for a maximum of $1,000,000$ update steps. We tokenized all data using the model's SentencePiece \cite{kudo-richardson-2018-sentencepiece} tokenizer.

To evaluate our models and to choose the best checkpoints, we used the BLEU score~\citep{papineni-etal-2002-bleu} calculated with the SacreBLEU~\citep{post-2018-call} implementation. In addition, we also evaluated the models using \textsc{chrF}~\citep{popovic2015chrf}. 
 
\subsubsection{Baseline models}
We train many-to-many (M2M) translation models by fine-tuning M2M-100 on the following subsets of the datasets described in \Cref{sec:dataset}. These include, the clean bitexts described in \Cref{sec:clean}, noisy bitext described in \Cref{sec:noisy}, and a mixture of the clean and noisy bitexts. The noisy bitext was only partially cleaned, as evidenced in \Cref{tab:data-heuristics}, using the heuristic rules mentioned in \Cref{sec:filter:heuristic}  without applying the proposed automatic filtering on data.

We trained these baseline models to compare and measure the efficacy of our filtering technique on the quality of the translation models. We submitted the model in (i) as our secondary system for this task.

\subsubsection{Models on filtered data only}
To evaluate the effect of the filtered data on the quality of the translation output, we train M2M models on the filtered data from the different models using a threshold of $0.5$ and $0.7$.

\subsubsection{Models on filtered and clean data}
We went further to train multilingual models on the concatenation of the noisy and clean text, and on the filtered and clean data for easier comparison. With this system, we were able to measure the amount of improvement we can obtain by including the clean bitext compared to training models only on the filtered bitext.

\section{Results and Discussion} \label{sec:result}

In \Cref{tab:flores,tab:mafand}, we report the BLEU and \textsc{chrF} scores obtained by the different models that we trained, as evaluated on the FLORES-101 devtest and MAFAND-MT test datasets, respectively.

\subsection{Baseline Models}
On average, the baseline model trained on the clean bitext performed impressively on the two evaluation datasets, despite the limited dataset size. On MAFAND-MT, the model trained on the clean bitext obtained a higher BLEU score than the model trained on the noisy bitext, and on FLORES-101, the reverse was true. This is likely due to the fact that the MAFAND-MT data is present in the clean bitext, and that the noisy bitext contains sentences that were taken from the web, including Wikipedia, which is the source of the FLORES-101 dataset. When we compared the model trained on the clean bitext to the model trained on the noisy bitext, we saw between a $+1$ and $+2$ improvement on FLORES-101 and between $+5$ and $+8$ improvement on MAFAND-MT for \texttt{lug}, \texttt{wol}, and \texttt{yor}. This confirms not only the importance of the data domain, but also the importance of data quality on the quality of the machine translation output.

After mixing the two datasets, the performance improved over using only the clean bitext by more than 6 BLEU on \texttt{hau}$\leftrightarrow$\texttt{eng}, and almost 3 BLEU on average across all languages on FLORES. The performance, though, was similar to using only the noisy bitext. On the MAFAND-MT test set, however, the performance deteriorated by almost 2 BLEU when compared to training on the clean bitext only. At language-pair level, \texttt{eng}$\rightarrow$\texttt{ibo} was affected more ($-9.14$ BLEU), followed by \texttt{eng}$\rightarrow$\texttt{wol}, whereas \texttt{yor}$\rightarrow$\texttt{eng} benefited tremendously ($+17.83$ BLEU). On average, training on the two bitexts marginally improves over using only the noisy bitext, and this is consistent on all the test sets.

Investigating the results in more depth, we found that the BLEU scores of the models are lower when translating into an African language, similar to the findings of \citet{adelani-etal-2022-thousand}. This effect is exacerbated for the languages with the fewest parallel sentences, such as \texttt{lug}, \texttt{wol}, and \texttt{yor}, except for \texttt{ibo}, which overall has the second-fewest parallel sentences, as shown in \autoref{tab:datastat}.

\begin{table*}
 \begin{center}
 \resizebox{\textwidth}{!}{
  \begin{tabular}{lrrrrrrrrrrrrrrrrr}\toprule
    \multirow{2}{*}{Models} & \multicolumn{7}{c}{\texttt{eng}$\rightarrow$x} & \multicolumn{1}{c}{\texttt{fra}$\rightarrow$x} & \multicolumn{7}{c}{x$\rightarrow$\texttt{\textit{eng}}} & \multicolumn{1}{c}{x$\rightarrow$\texttt{\textit{fra}}} & \multirow{2}{*}{Avg.} \\
    \cmidrule(lr){2-8}\cmidrule(lr){9-9}\cmidrule(lr){10-16}\cmidrule(lr){17-17}
     & \texttt{hau} & \texttt{ibo} & \texttt{lug} & \texttt{swa} & \texttt{tsn} & \texttt{yor} & \texttt{zul} & \texttt{wol} & \texttt{hau} & \texttt{ibo} & \texttt{lug} & \texttt{swa} & \texttt{tsn} & \texttt{yor} & \texttt{zul} & \texttt{wol} & \\ \midrule
    \multicolumn{18}{l}{\textbf{BLEU}} \\
    \midrule
    \multicolumn{18}{l}{\textbf{~~Baselines}} \\
    ~~Clean bitext & \textbf{9.00} & \textbf{20.83} & \textbf{11.67} & 25.81 & 18.64 & \textbf{9.86} & 14.50 & \textbf{8.91} & \textbf{12.49} & \textbf{19.24} & 20.00 & 29.28 & 20.44 & 16.98 & 23.20 & \textbf{7.77} & 16.79 \\
    ~~Noisy bitext & 5.24 & 10.37 & 6.12 & 25.35 & 16.61 & 3.61 & 15.23 & 0.98 & 8.52 & 12.83 & 14.35 & 28.37 & 21.34 & 13.14 & 26.74 & 1.57 & 13.15 \\
    ~~Clean $+$ Noisy bitext & 5.59 & 11.69 & 6.54 & 25.55 & 17.25 & 3.42 & 15.10 & 1.99 & 8.80 & 13.64 & 15.67 & 28.67 & 21.74 & \textbf{34.81} & 26.68 & 2.33 & 14.97 \\
    \midrule
    \multicolumn{18}{l}{\textbf{~~Filtered only}}  \\
    ~~albert-xlarge-0-7 & 7.75 & 16.33 & 7.56 & 26.45 & 23.01 & 4.59 & 17.63 & 0.86 & 9.93 & 15.59 & 16.77 & 30.92 & 30.92 & 16.46 & 29.47 & 3.09 & 16.08 \\
    \midrule
    \multicolumn{18}{l}{\textbf{~~Filtered $+$ Clean bitext}}  \\
    ~~albert-xlarge-0-5 & 8.49 & 18.16 & 10.11 & \textbf{27.89} & 22.99 & 5.37 & 17.68 & 5.46 & 11.73 & 17.53 & 20.63 & 32.38 & 27.07 & 17.84 & 29.88 & 5.52 & 17.42 \\
    ~~albert-xlarge-0-7 & 8.74 & 19.08 & 10.26 & 27.80 & \textbf{24.25} & 6.09 & \textbf{18.25} & 6.05 & 12.32 & 17.58 & \textbf{21.15} & \textbf{32.60} & \textbf{27.40} & 18.54 & \textbf{30.02} & 6.77 & \textbf{17.93} \\
    \midrule
    \multicolumn{18}{l}{\textbf{\textsc{chrF}}} \\
    \midrule
    \multicolumn{18}{l}{\textbf{~~Baselines}} \\
    ~~Clean bitext & 36.23 & 34.10 & \textbf{31.59} & 54.59 & 33.85 & \textbf{21.97} & 41.70 & \textbf{26.22} & 37.74 & 37.32 & 33.85 & 51.39 & 32.43 & 30.51 & 43.20 & 29.31 & 36.00 \\
    ~~Noisy bitext  & 40.24 & 31.27 & 25.84 & 59.14 & 38.88 & 19.18 & 46.98 & 8.90 & 44.80 & 38.71 & 34.58 & 56.25 & 40.57 & 33.28 & 49.26 & 19.15 & 36.69 \\
    ~~Clean $+$ Noisy bitext & 40.91 & 31.67 & 26.04 & 59.13 & 39.60 & 19.06 & 47.14 & 9.66 & 44.63 & 39.18 & 34.76 & 56.20 & 40.65 & 33.71 & 49.23 & 21.86 & 37.09 \\
    \midrule
    \multicolumn{18}{l}{\textbf{~~Filtered only}} \\
    ~~albert-xlarge-0-7 & \textbf{44.19} & 38.13 & 27.37 & 59.40 & 43.97 & 20.85 & \textbf{51.96} & 11.11 & 44.98 & 42.95 & 33.98 & 58.60 & \textbf{43.12} & 35.55 & \textbf{52.09} & 24.51 & 39.55 \\
    \midrule
    \multicolumn{18}{l}{\textbf{~~Filtered $+$ Clean bitext}} \\
    ~~albert-xlarge-0-5 & 43.38 & 37.88 & 29.70 & \textbf{61.47} & 43.30 & 20.57 & 51.06 & 18.73 & \textbf{45.53} & 42.77 & 36.14 & \textbf{58.93} & 43.11 & 36.61 & 52.06 & 28.26 & 40.59 \\
    ~~albert-xlarge-0-7 & 44.15 & \textbf{38.72} & 30.78 & 60.63 & \textbf{44.11} & 21.01 & 51.85 & 19.82 & 45.40 & \textbf{43.31} & \textbf{36.15} & 58.45 & 42.90 & \textbf{36.81} & 52.06 & \textbf{29.52} & \textbf{40.98} \\
    \bottomrule
    \end{tabular}
  }
  \vspace{-1mm}
  \caption{Performance of the multilingual model on the \textbf{MAFAND-MT} test set, with the maximum BLEU per column in \textbf{bold}. x represents African languages.}
  \label{tab:mafand}
\end{center}
\end{table*}

\subsection{Data Filtering Analysis}
We generally see that more filtering results in improved performance, corresponding to removing more noisy sentences from the data. Using less filtering, with a threshold of $0.5$, generally performed slightly worse than using a threshold of $0.7$. Both of these settings outperformed (a) using no filtering and (b) using no additional data.

We can also see the effect of the filtering steps on the training data in \Cref{tab:data-heuristics,tab:data-spc}. Filtering the data using heuristics resulted in only a small portion of the data being filtered out. Using the classifier, however,  caused a large amount of noisy data to be removed. When looking at the F1 scores of the classification models, we can see that ALBERT-xlarge has the lowest F1, followed by ALBERT-base and AfroXLMR-base. Looking at \Cref{tab:data-overlap}, we can see that ALBERT-xlarge is also the most strict filter, removing the most data, whereas AfroXLMR-base removes the least amount of data. Interestingly, the number of sentences marked as high-quality by all three models is surprisingly low, possibly indicating that these different models (particularly ALBERT-xlarge and AfroXLMR-base) focus on different features of the data.

Finally, we saw that a higher threshold resulted in improved translation performance, but ALBERT-xlarge (which is quite strict) had a lower F1 than the other models, possibly suggesting that F1 performance does not fully indicate the expected downstream performance on the actual translation task.

\subsubsection{The effect of filtering on translation models}
We fine-tune M2M-100 for multilingual translation on the filtered data, and as expected, our results (on average) demonstrate a considerable improvement when the translation model is trained on the filtered data rather than the original noisy texts. In particular, for many languages, training on the filtered data from ALBERT-xlarge with a threshold of $0.7$ outperformed the model trained on just the noisy bitext with at least a BLEU point.

Furthermore, we compared the performance of the model trained on only the clean data and on only the filtered data. Just as we saw with the baseline system, on MAFAND-MT, the model trained on the clean bitext performed better than the model trained on the filtered bitext, and on FLORES-101, the reverse was true. These results again confirm the importance of the filtering approach and further supports the observation that NMT engines are less robust to noise as found by \citet{khayrallah-koehn-2018-impact}, especially for low-resource settings.

\begin{table*}
 \begin{center}
 \resizebox{\textwidth}{!}{
  \begin{tabular}{lrrrrrrrrrrrrrrrrrrr}
    \toprule
    \multirow{2}{*}{Models} & \multicolumn{7}{c}{\texttt{eng}$\rightarrow$x} & \multicolumn{1}{c}{\texttt{fra}$\rightarrow$x} & \multicolumn{7}{c}{x$\rightarrow$\texttt{\textit{eng}}} & \multicolumn{1}{c}{x$\rightarrow$\texttt{\textit{fra}}} & \multirow{2}{*}{$x \to \text{afr}$} & \multirow{2}{*}{$\text{afr} \to x$} & \multirow{2}{*}{Avg.} \\
    \cmidrule(lr){2-8}\cmidrule(lr){9-9}\cmidrule(lr){10-16}\cmidrule(lr){17-17}
& \texttt{hau} & \texttt{ibo} & \texttt{lug} & \texttt{swa} & \texttt{tsn} & \texttt{yor} & \texttt{zul} & \texttt{wol} & \texttt{hau} & \texttt{ibo} & \texttt{lug} & \texttt{swa} & \texttt{tsn} & \texttt{yor} & \texttt{zul} & \texttt{wol} & & & \\
\midrule
\multicolumn{19}{l}{\textbf{BLEU}} \\
\midrule
~~Clean only & 10.7 & 11.9 & 4.5 & 24.3 & 10.1 & \textbf{4.2} & 6.0 & \textbf{4.4} & 15.7 & 15.0 & 12.2 & 27.5 & 9.7 & 8.8 & 18.5 & 7.1 & 9.5 & 14.3 & 11.9 \\
~~Filtered + Clean & \textbf{17.7} & \textbf{15.3} & \textbf{4.6} & \textbf{31.5} & \textbf{17.8} & 3.2 & \textbf{11.1} & 1.5 & \textbf{22.7} & \textbf{20.9} & \textbf{15} & \textbf{35.2} & \textbf{21.2} & \textbf{14.2} & \textbf{26.8} & \textbf{7.6} & \textbf{12.8} & \textbf{20.4} & \textbf{16.6} \\
\midrule
\multicolumn{19}{l}{\textbf{\textsc{chrF}2++}} \\
\midrule
~~Clean only & 36.0 & 34.6 & \textbf{29.0} & 52.2 & 33.8 & \textbf{21.8} & 36.3 & \textbf{25.4} & 38.0 & 38.2 & 33.4 & 50.4 & 31.6 & 29.4 & 41.6 & \textbf{28.0} & 33.6 & 36.3 & 35.0 \\
~~Filtered + Clean & \textbf{43.4} & \textbf{38.6} & 27.2 & \textbf{57.7} & \textbf{41.9} & 19.4 & \textbf{44.8} & 17.9 & \textbf{45.2} & \textbf{44.6} & \textbf{35.4} & \textbf{57.1} & \textbf{43.6} & \textbf{35.3} & \textbf{49.1} & 27.7 & \textbf{36.4} & \textbf{42.2} & \textbf{39.4} \\
    \bottomrule
  \end{tabular}
  }
  \vspace{-1mm}
  \caption{Performance of the submitted models on the wmt22 test sets as provided by the organizers. We submitted two models. The primary one, denoted \textit{Filtered + Clean}, was trained on the clean bitext as well as the data filtered by ALBERT-xlarge with a threshold of $0.7$. The secondary (or contrastive) approach, denoted \textit{Clean only}, was trained only on the clean bitext. The $x \to \text{afr}$ and $\text{afr} \to x$ columns contain the average performance for translations to and from African languages, respectively. \textit{avg} contains the average over all language pairs.}
  \label{tab:wmt22}
\end{center}
\end{table*}
 
\subsection{Clean vs. filtered data}
We find that on FLORES-101, adding in noisy, unfiltered data improves the results over just using the true parallel data. On MAFAND-MT, however, it generally reduces the BLEU score significantly. For both datasets, adding appropriately filtered data results in the highest performance averaged over all the languages, although for some specific languages, just using true parallel data resulted in the best performance.

Our performance on the test set provided by the organizers~\citep{adelani-etal-2022-findings} is shown in \autoref{tab:wmt22}. Here we can see that our primary model, which was trained on the clean bitext as well as the filtered data (filtered using ALBERT-xlarge, $t=0.7$), significantly outperforms the model trained only on the clean bitext. We also see that our approach seems to have a larger performance gain when translating \textit{from} African languages compared to translating \textit{to} them.

\section{Conclusion and Future Work} \label{sec:conclude}
In this work, we used a sentence-pair classifier to classify parallel data as being aligned, or not. Using this approach, we filtered out a large portion of the original, noisy, data and fine-tuned existing large language models on this new data. Our results show that training on the filtered data significantly increases the performance of the models, resulting in improved translations. In particular, our approach outperforms (i) training only on clean data, (ii) training only on filtered data, and (iii) training on the original dataset, consisting of clean and noisy data. This provides additional evidence in favor of prioritizing data quality over quantity, as well as the need for more advanced noise detection and filtering tools.
There are numerous potential avenues for future work; one option is to use a multilingual model as the sentence classifier instead of using a separate model per language, to leverage commonalities between different languages~\citep{adelani2021MasakhaNER,conneau2019Unsupervised}. Secondly, a more in-depth study of the effect of the threshold parameter on the final BLEU score would be useful. We would also like to understand the reasons behind the performance by analyzing the filtered data more in depth. Finally, given more computational resources, we will (i)~train the classifier for more epochs, using other language models and/or using different quality thresholds, (ii)~use longer sentence length than the current 128, (iii)~train the translation models on AfroXLMR and ALBERT-base filtered data, and (iv)~use the filtering approach on more languages, to evaluate its generalizability.
Ultimately, we hope that this filtering approach could lead to the use of cleaner data to train translation models, improving the overall translation quality for low-resourced languages.

\section*{Acknowledgements}
Idris Abdulmumin acknowledges funding from Google through the Africa Travel Grants. Computations were performed using High Performance Computing infrastructure provided by the Mathematical Sciences Support unit at the University of the Witwatersrand, LIAAD~INESC TEC, Portugal, and DFKI GmbH. We acknowledge support from the Machine Learning Group, EISLAB, Luleå University of Science and Technology, Sweden. In addition, we thank the organizers for the compute grant, which allowed us to run some of our experiments. Finally, we thank Dr. Rachel Bawden, and the anonymous reviewers of the 2022 WMT shared task for their helpful feedback.
\bibliography{anthology,acl2020,custom}
\bibliographystyle{acl_natbib}

\appendix
\begin{table*}[t]
    \scalebox{0.95}{
    \small
    \begin{tabular}{@{}lrrrrrrr|r@{}}
    \hline
        \multirow{2}{*}{Data} & \multicolumn{7}{c}{en} & fr \\ \cline{2-9}
        & \texttt{hau} & \texttt{ibo} & \texttt{lug} & \texttt{swa} & \texttt{tsn} & \texttt{yor} & \texttt{zul} & \texttt{wol} \\ \hline
        \multicolumn{9}{l}{True Parallel} \\ \hline
        MAFAND-MT & $3,098$ & $6,998$ & $4,075$ & $30,782$ & $2,100$ & $6,644$ & $3,500$ & $3,360$ \\
        Tanzil & $128,376$ & - & - & $138,253$ & - & - & - & - \\
        GlobalVoices & - & - & - & $32,307$ & - & $137$ & - & - \\
        tico-19 & $3,071$ & - & $3,071$ & $3,071$ & - & - & $3,071$ & - \\
        ELRC\textunderscore2922 & - & - & - & $607$ & - & - & - & - \\
        EUbookshop & - & - & - & $18$ & - & - & - & - \\
        Tatoeba & $57$ & $22$ & $3$ & $395$ & $31$ & $37$ & $70$ & $67$ \\
        bible-uedin & - & - & - & - & - & - & $15,907$ & $7,918$ \\
        QED & $124$ & $12$ & $740$ & $18,192$ & - & $52$ & $1,624$ & $66$ \\
        Mozilla-I10n & $4,952$ & $4,172$ & $5,931$ & $7,798$ & - & $4,095$ & - & $7,041$ \\ \hline
        \textbf{Total (TP)} & $139,678$ & $11,204$ & $13,820$ & $231,423$ & $2,131$ & $10,965$ & $24,172$ & $18,452$ \\ \hline\hline
        \multicolumn{9}{l}{Automatcally Aligned} \\ \hline
        WMT22 African & $2,309,758$ & $172,973$ & $3,450,573$  & $23,358,739$ & $5,931,529$ & $1,455,571$ & $3,862,020$ & $189,659$ \\
        WebCrawl Afr. & $16,950$ & $3,372$ & $10,809$ & $193,518$ & $77,976$ & $18,924$ & $152,724$ & - \\
        LAVA Corpus & - & - & $20,993$ & $371,864$ & - & - & - & - \\
        WikiMatrix & - & - & - & $51,387$ & - & - & - & - \\
        CCAligned & $339,178$ & $148,147$ & $14,702$ & $2,044,993$ & $71,254$ & $175,193$ & $126,103$ & $-$ \\
        CCMatrix & $5,861,080$ & $80,385$ & - & $5,756,664$ & - & - & - & - \\
        ParaCrawl & - & - & - & $132,521$ & - & - & - & - \\
        GNOME & $5,466$ & $23,767$ & $4,578$ & $40$ & - & $10,234$ & $44,605$ & - \\
        KDE4 & $1,493$ & - & - & - & - & - & - & - \\
        TED2020 & $27$ & $210$ & - & $9,745$ & - & - & - & - \\
        XLEnt & $436,602$ & $69,820$ & $1,054$ & $871,902$ & $4,781$ & $51,173$ & $28,394$ & $4,082$ \\
        Ubuntu & $242$ & $635$ & $637$ & $986$ & - & $141$ & $4,718$ & $220$ \\
        wikimedia & $23,385$ & $12,279$ & $1,315$ & $3,765$ & $969$ & $8,521$ & $1,226$ & $679$ \\
        MultiCCAligned & - & - & - & - & - & - & - & $24,256$ \\ \hline
        \textbf{Total (AA)} & $8,994,181$ & $511,588$ & $3,504,661$ & $32,796,124$ & $6,086,509$ & $1,719,757$ & $4,219,790$ & $218,896$ \\ \hline\hline
        \textbf{Total (ALL)} & $9,133,859$ & $522,792$ & $3,518,481$ & $33,027,547$ & $6,088,640$ & $1,730,722$ & $4,243,962$ & $237,348$ \\ \hline
    \end{tabular}
    }
    \caption{Training Data Used --- TP=True Parallel; AA=Automatically Aligned}
    \label{tab:datastat}
\end{table*}

\section{Appendix - Data Sources}
\label{app:data-sources}
Datasets used in this project and their sources, as listed in \Cref{tab:datastat}: MAFAND-MT, wmt22\_african, LAVA Corpus,\footnote{\url{https://drive.google.com/drive/folders/179AkJ0P3fZMFS0rIyEBBDZ-WICs2wpWU}} XLEnt, Tanzil, WikiMatrix, CCAligned, CCMatrix, GlobalVoices,\footnote{\url{https://casmacat.eu/corpus/global-voices.html}}$^,$\footnote{\url{https://globalvoices.org/}} ParaCrawl,\footnote{\url{https://paracrawl.eu/}} GNOME,\footnote{\url{https://l10n.gnome.org/}} tico-19,\footnote{\url{https://tico-19.github.io/index.html}} ELRC\_2922,\footnote{\url{https://elrc-share.eu/repository/browse/covid-19-health-wikipedia-dataset-multilingual-53-en-x-language-pairs/fe23e2c28c8311ea913100155d0267066f62c6b30ac0429f8d497df0abd2ef72/}} EUbookshop,\footnote{\url{http://bookshop.europa.eu}} KDE4,\footnote{\url{http://www.lt-innovate.org/lt-observe/resources/kde4-kde4-localization-files-v2}} TED2020, Tatoeba,\footnote{\url{https://tatoeba.org/en/}} Ubuntu,\footnote{\url{https://translations.launchpad.net/}} bible-uedin, wikimedia,\footnote{\url{https://dumps.wikimedia.org/other/contenttranslation/}} QED, MultiCCAligned and Mozilla-I10n.

\end{document}